\documentclass[10pt,twocolumn,letterpaper]{article}

\usepackage{cvpr}              % To produce the CAMERA-READY version
\usepackage{graphicx}
\usepackage{amsmath}
\usepackage{amssymb}
\usepackage{booktabs}
\usepackage{multirow}

\usepackage{algorithm}
\usepackage{algorithmic}

\usepackage[pagebackref,breaklinks,colorlinks]{hyperref}

\usepackage[capitalize]{cleveref}
\crefname{section}{Sec.}{Secs.}
\Crefname{section}{Section}{Sections}
\Crefname{table}{Table}{Tables}
\crefname{table}{Tab.}{Tabs.}

\def\cvprPaperID{5452} % *** Enter the CVPR Paper ID here
\def\confName{CVPR}
\def\confYear{2023}

\begin{document}

\title{Temporal Attention Unit: Towards Efficient Spatiotemporal Predictive Learning}

% \and
\author{Cheng Tan$^{*}$, Zhangyang Gao\footnotemark[1]\thanks{Equal contribution.}, Lirong Wu, Yongjie Xu, Jun Xia, Siyuan Li, Stan Z. Li\thanks{Corresponding author.} \\ %$\dagger$
% $^{1}$ Zhejiang University
AI Lab, Research Center for Industries of the Future, Westlake University \\
{\tt\small \{tancheng,gaozhangyang,wulirong,xuyongjie,xiajun,lisiyuan,stan.zq.li\}@westlake.edu.cn}
}
\maketitle
%%%%%%%%% ABSTRACT
% abstract和conclusion要好好写，第一印象很重要！！！
% abstract和conclusion要好好写，第一印象很重要！！！
% abstract和conclusion要好好写，第一印象很重要！！！
% The problem: limitation of existing methods or lack of existing studies
% How we solve problem (briefly)
% Our method in one more sentence
% Empirical results
% Maybe highlight the contribution (To our knowledge, we are the first to ...)
% abstract和conclusion要好好写，第一印象很重要！！！
% abstract和conclusion要好好写，第一印象很重要！！！
% abstract和conclusion要好好写，第一印象很重要！！！
\begin{abstract}
Spatiotemporal predictive learning aims to generate future frames by learning from historical frames. In this paper, we investigate existing methods and present a general framework of spatiotemporal predictive learning, in which the spatial encoder and decoder capture intra-frame features and the middle temporal module catches inter-frame correlations. While the mainstream methods employ recurrent units to capture long-term temporal dependencies, they suffer from low computational efficiency due to their unparallelizable architectures. To parallelize the temporal module, we propose the Temporal Attention Unit (TAU), which decomposes temporal attention into intra-frame statical attention and inter-frame dynamical attention. Moreover, while the mean squared error loss focuses on intra-frame errors, we introduce a novel differential divergence regularization to take inter-frame variations into account. Extensive experiments demonstrate that the proposed method enables the derived model to achieve competitive performance on various spatiotemporal prediction benchmarks.
\end{abstract}

%%%%%%%%% BODY TEXT
\section{Introduction}
\label{sec:intro}
% 监督学习 -> 无监督学习
The last decade has witnessed revolutionary advances in deep learning across various supervised learning tasks such as image classification~\cite{he2016deep,tan2019efficientnet,li2022efficient}, object detection~\cite{ren2015faster,redmon2016you}, computational biology~\cite{jumper2021highly,tan2022target,tan2022rfold,gao2022alphadesign,gao2022pifold}, and etc. Despite significant breakthroughs in supervised learning, which relies on large-scale labeled datasets, the potential of unsupervised learning remains largely untapped. 
% 无监督学习 -> 自监督学习 -> 时空预测学习
Self-supervised learning that designs pretext tasks to produce labels derived from the data itself is recognized as a subset of unsupervised learning. In the context of self-supervised learning, \textit{contrastive self-supervised learning}~\cite{he2020momentum,chen2020simple,NEURIPS2020_f3ada80d,chen2020big,wang2020understanding,zbontar2021barlow,tan2021co,tan2022hyperspherical} predicts the noise contrastive estimation from predefined positive or negative pairs, and \textit{masked self-supervised learning}~\cite{kenton2019bert,yang2019xlnet,lewis2020bart,dosovitskiy2020image,he2021masked,liu2021swin} predicts the masked patches from the visible patches. Unlike these image-level self-supervised learning, \textit{spatiotemporal predictive learning} that predicts future frames from past frames at the video-level~\cite{finn2016unsupervised,locatello2019challenging,greff2019multi,mathieu2019disentangling,khemakhem2020variational,castrejon2019improved}. 

\begin{figure}[htbp]
\centering
\includegraphics[width=0.48\textwidth]{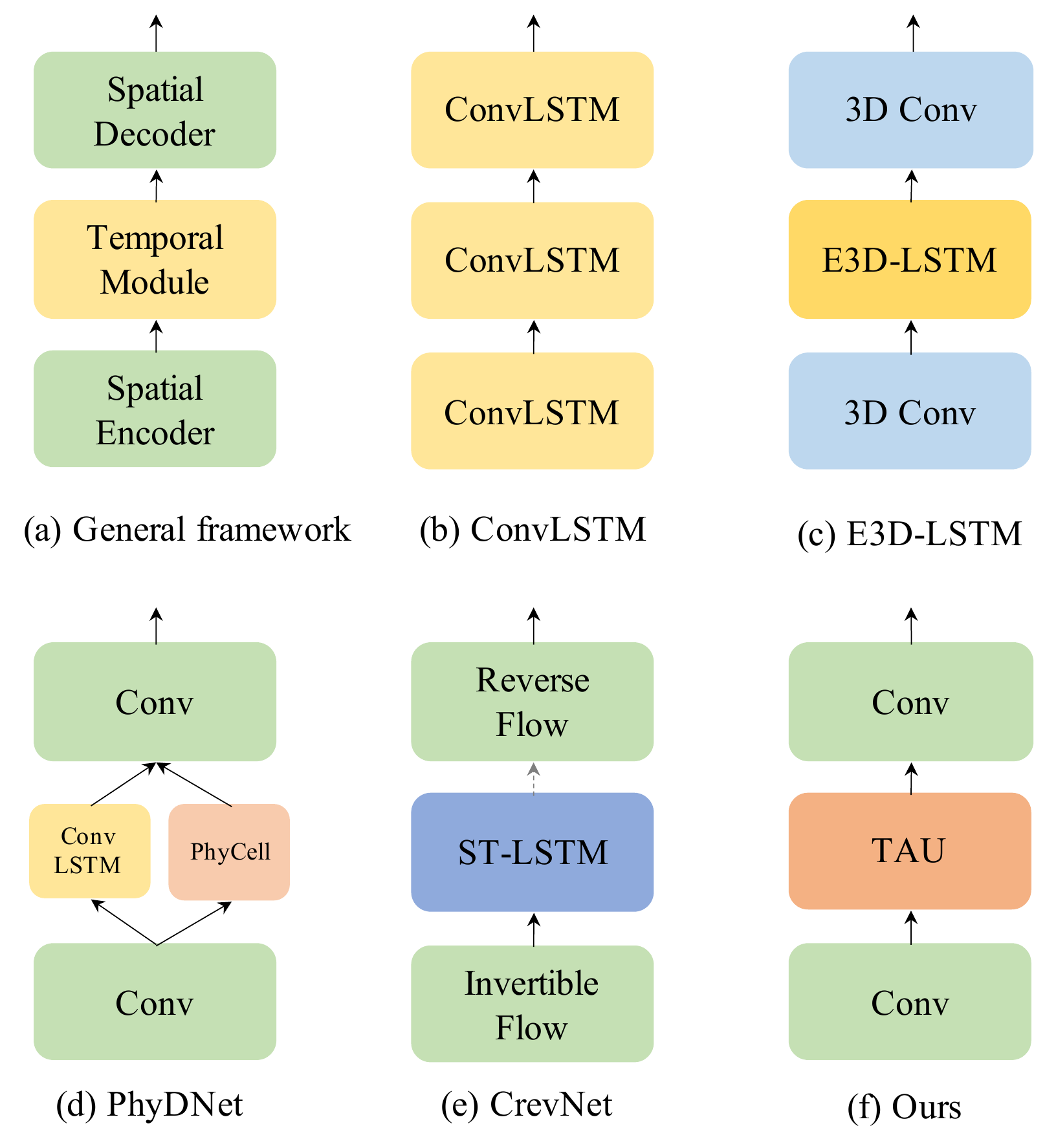} 
\caption{Comparison of model architectures among common spatiotemporal predictive learning methods. Note that we denote 2D convolutional neural networks as Conv while 3D Conv means 3D convolutional neural networks.}
\label{fig:general_architecture} 
\end{figure}

% 时空预测学习 -> 视频预测的重要性
Accurate spatiotemporal predictive learning can benefit broad practical applications in climate change~\cite{shi2015convolutional,reichstein2019deep}, human motion forecasting~\cite{zhang2017learning,wang2018rgb}, traffic flow prediction~\cite{fang2019gstnet,wang2019memory}, and representation learning~\cite{qian2021spatiotemporal,jenni2020video}. The significance of spatiotemporal predictive learning primarily lies in its potential of exploring both spatial correlation and temporal evolution in the physical world. Moreover, the self-supervised nature of spatiotemporal predictive learning aligns well with human learning styles without a large amount of labeled data. Massive videos can provide a rich source of visual information, enabling spatiotemporal predictive learning to serve as a generative pre-training strategy~\cite{locatello2019challenging,hinton2021represent} for feature representation learning towards diverse downstream visual supervised tasks.

% 总结现有方法，提出general framework（配图）
Most of the existing methods~\cite{convlstm,predrnn,predrnn++,e3dlstm,mim,phydnet,predrnnv2,crevnet, finn2016unsupervised,shi2017deep,su2020convolutional,kalchbrenner2017video,bhagat2020disentangling,wu2021motionrnn,gehring2017convolutional,villegas2018hierarchical,villegas2019high,minderer2019unsupervised,denton2018stochastic,denton2017unsupervised,cheng2021graph,babaeizadeh2021fitvid,jin2020exploring} in spatiotemporal predictive learning employ hybrid architectures of convolutional neural networks and recurrent units in which spatial correlation and time evolution can be learned, respectively. Inspired by the success of long short-term memory (LSTM)~\cite{hochreiter1997long} in sequential modeling, ConvLSTM~\cite{convlstm} is a seminal work on the topic of spatiotemporal predictive learning that extends fully connected LSTM to convolutional LSTM towards accurate precipitation nowcasting. PredRNN~\cite{predrnn} is an admirable work that proposes Spatiotemporal LSTM (ST-LSTM) units to model spatial appearances and temporal variations in a unified memory pool. This work provides insights on designing typical recurrent units for spatiotemporal predictive learning and inspires a series of subsequent works~\cite{predrnn++, byeon2018contextvp, mim, predrnnv2}. E3D-LSTM~\cite{e3dlstm} integrates 3D convolutional neural networks into recurrent units towards good representations with both short-term frame dependencies and long-term high-level relations. PhyDNet~\cite{phydnet} introduces a two-branch architecture that involves physical-based PhyCells and ConvLSTMs for performing partial differential equation constraints in latent space. CrevNet~\cite{crevnet} proposes an invertible two-way autoencoder based on flow~\cite{dinh2014nice,dinh2016density_realnvp} and a conditionally reversible architecture for spatiotemporal predictive learning. As shown in Figure~\ref{fig:general_architecture} (a), we present a general framework consisting of the spatial encoder/decoder and the middle temporal module, abstracted from these methods. Though these spatiotemporal predictive learning methods have different temporal modules and spatial encoders/decoders, they basically share a similar framework.

% 总结general network下其他方法的问题，然后引出自己的方法。
Based on the general framework, we argue that the temporal module plays an essential role in spatiotemporal predictive learning. While the common choice of the temporal module is the recurrent-based units, we explore a novel parallelizable attention module named Temporal Attention Unit (TAU) to capture time evolution. The proposed temporal attention is decomposed into intra-frame statical attention and inter-frame dynamical attention. Furthermore, we argue that the mean square error loss only focuses on intra-frame differences, and we propose a differential divergence regularization that also cares about the inter-frame variations. Keeping the spatial encoder and decoder as simple as 2D convolutional neural networks, we deliberately implement our proposed TAU modules and surprisingly find the derived model achieves competitive performance as those recurrent-based models. This observation provides a new perspective to improve spatiotemporal predictive learning by parallelizable attention networks instead of common-used recurrent units. 

% Experiments
We conduct experiments on various datasets with different experimental settings: (1) Standard spatiotemporal predictive learning. Quantitative results on various datasets demonstrate our proposed method can achieve competitive performance on standard spatiotemporal predictive learning. (2) Generalization ability. To verify the generalization ability, we train our model on KITTI and test it on the Caltech Pedestrian dataset with different domains. (3) Predict future frames with flexible lengths. We tackle the long-length frames by feeding the predicted frames as the input and find the performance is consistently well. Through the superior performance in the above three experimental settings, we demonstrate that our proposed model can provide a novel manner that learns temporal dependencies without recurrent units.

\section{Related works}

\subsection{Self-supervised learning}
Deep learning has been well developed and applied in various fields~\cite{liu2022decoupled,cao2022survey,liu2022automix,zheng2022using,zheng2021enhancing}. Learning from massive data enables tremendous progress in supervised learning. By designing pretext tasks and generating labels from the data itself, self-supervised learning obtains supervisory signals. The model learns valuable representations by solving pretext tasks that leverage the underlying structure of the data. Early works on visual self-supervised learning design pretext tasks such as colorization~\cite{zhang2016colorful}, inpainting~\cite{pathak2016context}, rotation~\cite{gidaris2018unsupervised}, jigsaw~\cite{noroozi2016unsupervised}. Contrastive self-supervised learning~\cite{he2020momentum,chen2020simple,NEURIPS2020_f3ada80d,chen2020big,wang2020understanding,zbontar2021barlow} is a dominant manner in visual self-supervised learning that aims at a pretext task of grouping similar samples closer and diverse samples away from each other. However, contrastive self-supervised learning is limited by making pairs by multiple images, which affects its ability on small-scale datasets. Masked self-supervised learning~\cite{kenton2019bert,yang2019xlnet,lewis2020bart,dosovitskiy2020image,he2021masked,liu2021swin}, which predicts the masked patches from the visible ones, is another research direction. Although masked pretraining has great success in natural language processing, its applications in visual tasks are challenging. Spatiotemporal predictive learning is another promising branch of self-supervised learning that focus on video-level information and predicts future frames conditioned on past frames. 

In contrast to the above image-level methods, spatiotemporal predictive learning focus on video-level information and predicts future frames conditioned on past frames. By learning the intrinsic motion dynamics, the model is enabled to easily decouple the foreground and background.

\subsection{Spatiotemporal predictive learning}

Recurrent models have achieved remarkable advances in spatiotemporal predictive learning. Inspired by recurrent neural networks, VideoModeling~\cite{marc2014video} adopts language modeling and quantizes the image patches into a large dictionary for recurrent units. ConvLSTM~\cite{convlstm} leverages convolutional neural networks to model the LSTM architecture. PredNet~\cite{prednet} persistently predicts future video frames using deep recurrent convolutional neural networks with bottom-up and top-down connections. PredRNN~\cite{predrnn} proposes a Spatiotemporal LSTM unit that extracts and memorizes spatial and temporal representations simultaneously, and its following work PredRNN++~\cite{predrnn++} further introduces gradient highway unit and Casual LSTM to adaptively capture temporal dependencies. E3D-LSTM~\cite{e3dlstm} designs eidetic memory transition in recurrent convolutional units. Conv-TT-LSTM~\cite{su2020convolutional} employs a higher-order ConvLSTM to predict by combining convolutional features across time. MotionRNN~\cite{wu2021motionrnn} focuses on motion trends and transient variations. LMC-Memory~\cite{lee2021video} introduces a long-term motion context memory using memory alignment learning. PredRNN-v2~\cite{predrnnv2} extends PredRNN by leveraging a memory decoupling loss and curriculum learning strategy.

% 尽管recurrent取得了不错的效果，但他们suffer from...。因此我们提出了...
% Although these recurrent-based methods have achieved great success in spatiotemporal predictive learning, they suffer from complex computational efficiency. To parallelize the modeling of temporal evolution, we propose TAU, which employs visual attention mechanism without the recurrent architecture.
Instead of using recurrent-based methods that are computationally expensive for spatiotemporal predictive learning, we introduce TAU, a model that uses visual attention mechanism to parallelize the temporal evolution without the recurrent structure.
% There are prior arts that have some similarity with our proposed model...但是区别在哪里...
There are prior arts that have some similarities with our proposed model. PredCNN~\cite{predcnn} and TrajectoryCNN~\cite{liu2020trajectorycnn} implement pure convolutional neural networks as the temporal module. SimVP~\cite{Gao_2022_CVPR} is a seminal work that applies blocks of Inception modules with a UNet architecture to learn the temporal evolution. Though their temporal modules are parallelizable, we argue that convolutions alone cannot capture long-term dependencies. Moreover, SimVP provides a simple baseline with minor complex attachment but a large space for further improvements. In general, SimVP first downsamples video sequences to reduce the computation, then uses Inception-UNet to learn essential spatiotemporal relationships, and upsamples the representations to predict future frames. Our work aims to replace the pivotal Inception-UNet with efficient attention modules that promote prediction performance. In this paper, we employ a simple yet effective attention mechanism to enable the temporal module not only to be parallelizable but also to capture long-term time evolution.

\section{Methods} % 1.5 - 2.5 pages

\subsection{Preliminaries}

We formally define the spatiotemporal predictive learning problem as follows. Given a video sequence $\mathcal{X}^{t, T} = \{\mathbf{x}^i\}_{t-T+1}^t$ at time $t$ with the past $T$ frames, we aim to predict the subsequent $T'$ frames $\mathcal{Y}^{t+1, T'} = \{\mathbf{x}^{i}\}_{t+1}^{t+1+T'}$ from time $t+1$, where $\mathbf{x}_i \in \mathbb{R}^{C \times H \times W}$ is usually an image with channels $C$, height $H$, and width $W$. In practice, we represent the video sequences as tensors, i.e.,  $\mathcal{X}^{t, T} \in \mathbb{R}^{T \times C \times H \times W}$ and $\mathcal{Y}^{t+1, T'} \in \mathbb{R}^{T' \times C \times H \times W}$.

The model with learnable parameters $\Theta$ learns a mapping $\mathcal{F}_\Theta: \mathcal{X}^{t, T} \mapsto \mathcal{Y}^{t+1, T'}$ by exploring both spatial and temporal dependencies. In our case, the mapping $\mathcal{F}_\Theta$ is a neural network model trained to minimize the difference between the predicted future frames and the ground-truth future frames. The optimal parameters $\Theta^*$ are:
\begin{equation}
  \Theta^* = \arg\min_{\Theta} \mathcal{L}(\mathcal{F}_\Theta(\mathcal{X}^{t, T}), \mathcal{Y}^{t+1, T'}),
\end{equation}
where $\mathcal{L}$ is a loss function that evaluates such differences.

\subsection{Overview}

\begin{figure}[htbp]
\centering
\includegraphics[width=0.44\textwidth]{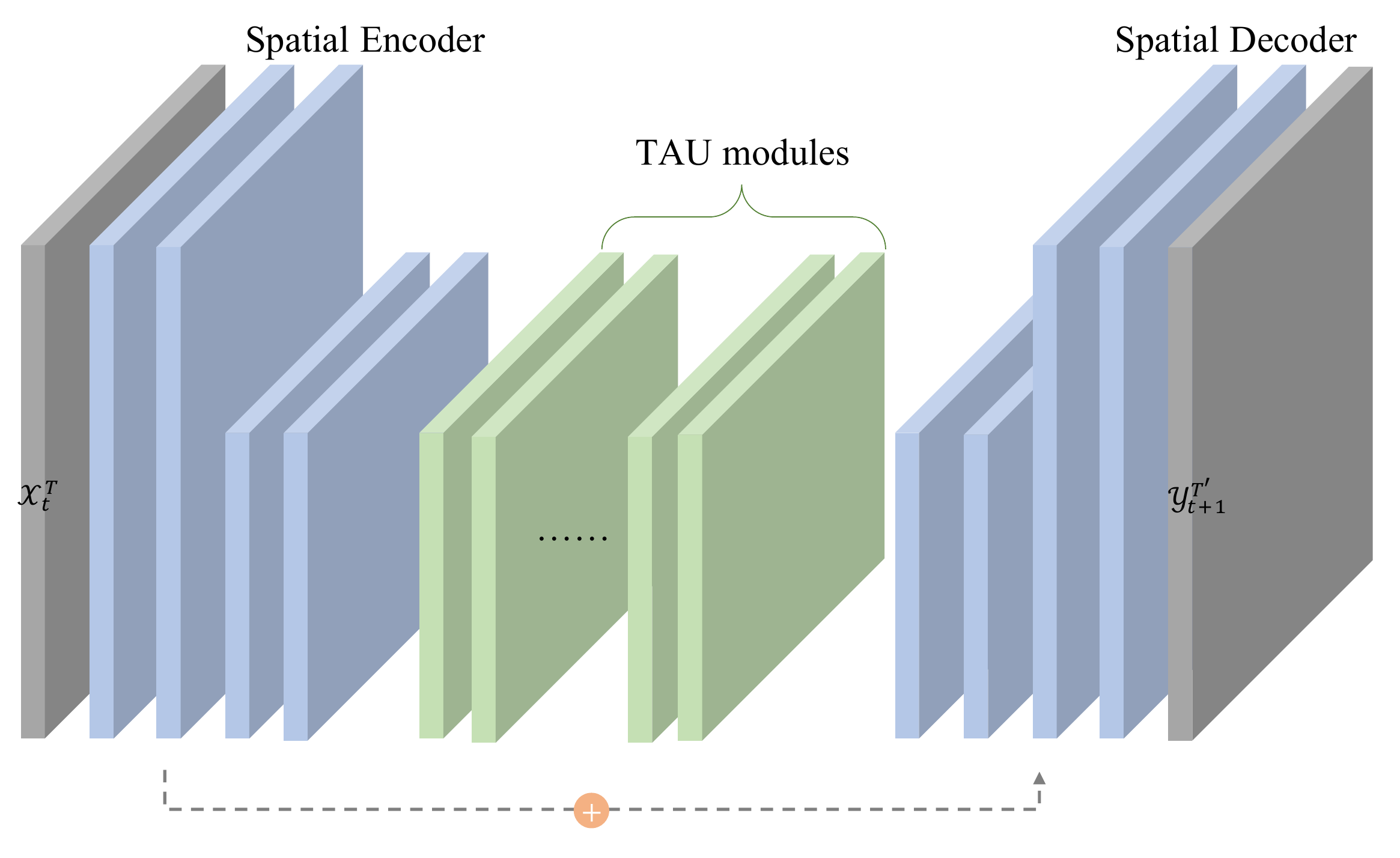} 
\caption{The overview architecture of our proposed model.}
\label{fig:overview} 
\end{figure}

We illustrate the overview model in Figure~\ref{fig:overview} using the input Moving MNIST~\cite{srivastava2015unsupervised} data as an example.

Striving for simplicity, the model follows the general framework in Figure~\ref{fig:general_architecture}, while the spatial encoder consists of four vanilla 2D convolutional layers, and the spatial decoder consists of four 2D transposed convolutional layers ('Conv2d' and 'ConvTranspose2d' in PyTorch respectively). We add a residual connection from the first convolutional layer to the last transposed convolutional layer for preserving the spatial-dependent features. Stacks of TAU modules are in the middle of the spatial encoder and decoder to extract temporal-dependent features. Though our model is simple, it can efficiently learn both spatial-dependent and temporal-dependent features without recurrent architectures.

\subsection{Temporal Attention Unit}

Suppose a batch of input video tensors $\mathcal{B} \in \mathbb{R}^{B \times T \times C \times H \times W}$ with the number of videos $B = |\mathcal{B}|$ is given. In the spatial encoder and decoder, we reshape the sequential input data $B \times T \times C \times H \times W$ as $(B \times T) \times C \times H \times W$ so that only spatial correlations are taken into account. In the temporal module, we reshape the feature $B \times T \times C \times H \times W$ as $B \times (T \times C) \times H \times W$ so that frames are arranged in order on the channel dimension.

\begin{figure}[htbp]
\centering
\includegraphics[width=0.48\textwidth]{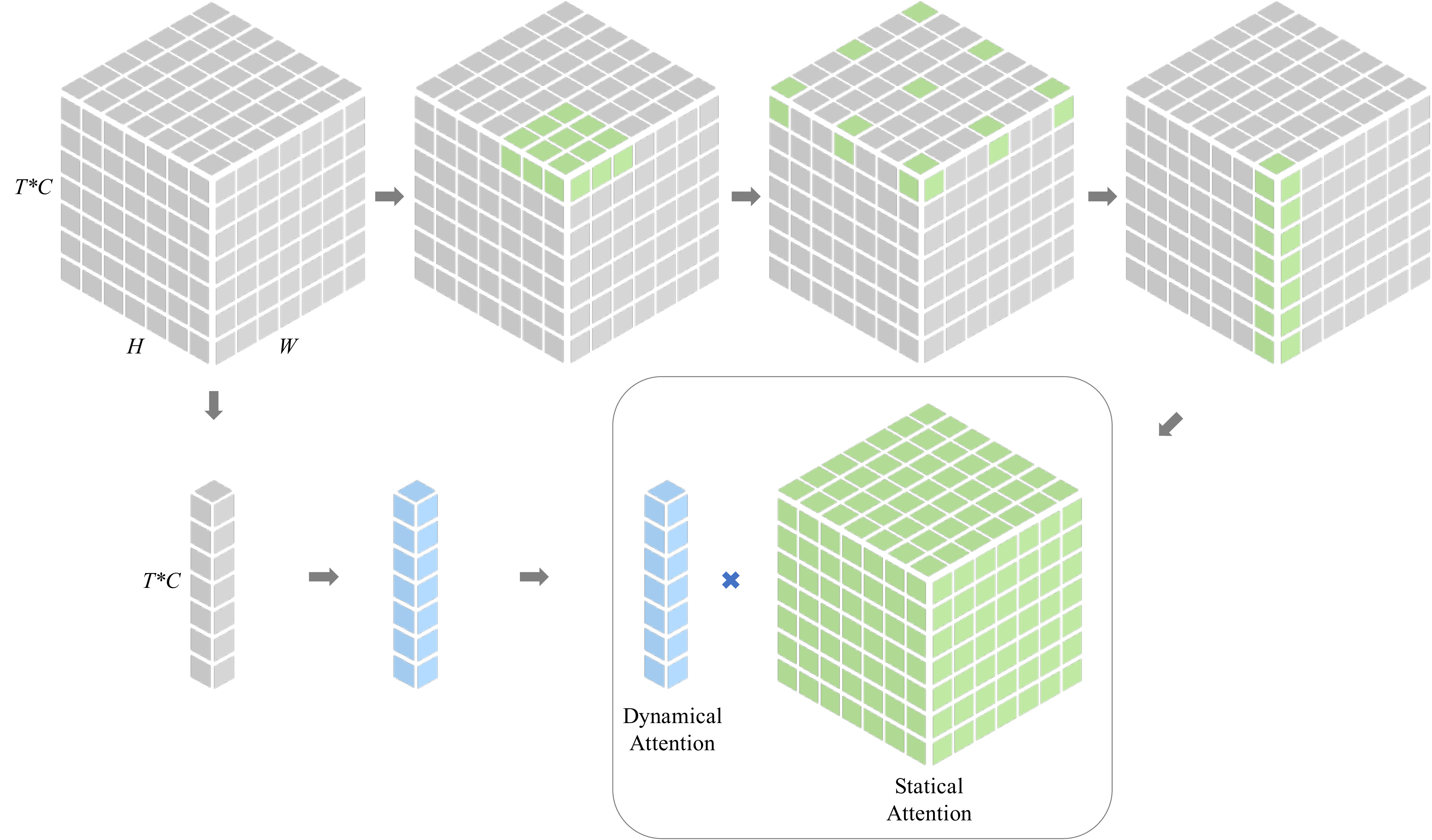} 
\caption{The intra-frame statical attention and the inter-frame dynamical attention.}
\label{fig:tau} 
\end{figure}

We decompose the temporal attention into the intra-frame statical attention and the inter-frame dynamical attention, as shown in Figure~\ref{fig:tau}. Inspired by the recent progress of vision Transformers (ViTs)~\cite{dosovitskiy2020image,liu2021swin} and large kernel convolutions~\cite{liu2022convnet,ding2022scaling,guo2022visual}, we propose to employ small kernel depth-wise convolutions (DW Conv), depth-wise convolutions with dilations (DW-D Conv), and 1$\times$1 convolutions to model the large kernel convolutions. Through the obtained large receptive field on intra-frames, the statical attention is able to capture long-range dependencies. However, the statical attention alone is not enough for learning temporal evolutions along the timeline. Thus, we employ the dynamical attention that learns the attention weights of channels in a squeeze-and-excitation manner~\cite{hu2018squeeze}. The final attention is the product of dynamical attention and statical attention.

\begin{figure}[htbp]
\centering
\includegraphics[width=0.46\textwidth]{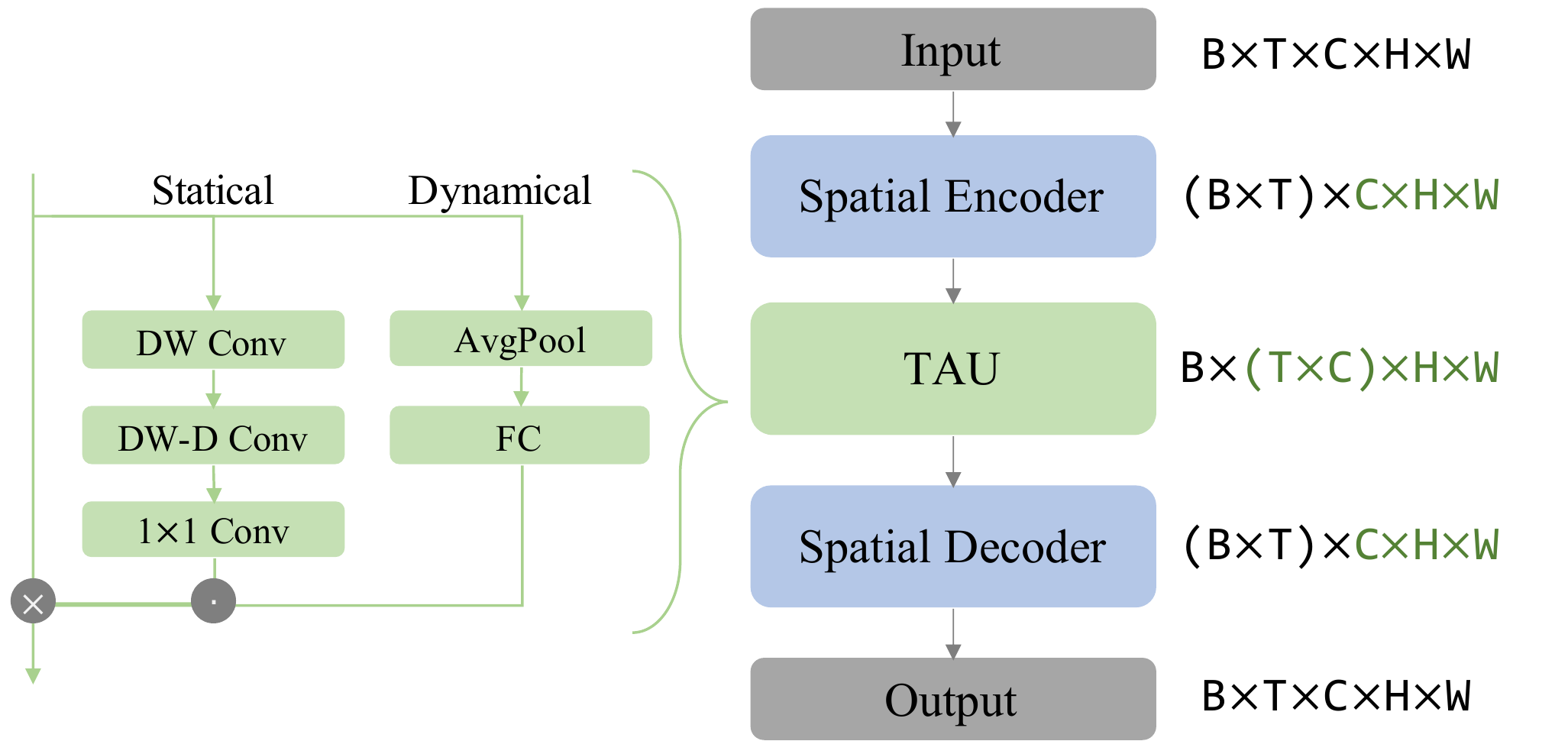} 
\caption{The detailed schema of our model.}
\label{fig:flow_chart} 
% \vspace{-3mm}
\end{figure}

We show the detailed scheme of our model in Figure~\ref{fig:flow_chart}. The proposed TAU module can be formally expressed as:
\begin{equation}
\begin{aligned}
\text{SA} &= \text{Conv}_{1\times 1}(\text{DW-D Conv}(\text{DW Conv}(H))), \\
\text{DA} &= \text{FC}(\text{AvgPool}(H)), \\
H' &= (\text{SA} \; \otimes \; \text{DA}) \; \odot \; H,
\end{aligned}
\end{equation}
where $H \in \mathbb{R}^{B \times (T \times C') \times H \times W}$ is the hidden feature that will be fed into the TAU module, $\text{SA} \in \mathbb{R}^{B\times (T \times C') \times H \times W}, \text{DA} \in \mathbb{R}^{B\times (T\times C') \times 1 \times 1}$ denote the statical and dynamical attention, $\text{FC}$ and $\text{AvgPool}$ are fully connected layers and the average pooling. We represent the Kronecker product by $\otimes$ and the Hadamard product by $\odot$.

\subsection{Differential Divergence Regularization}

To improve the prediction of our model, we further propose a differential divergence regularization that forces the model to learn the differences between consecutive frames and be aware of the inherent variation. 

Given the predicted frames $\widehat{\mathcal{Y}} = \mathcal{F}_\Theta(\mathcal{X}) \in \mathbb{R}^{T' \times C \times H \times W}$ and its corresponding ground-truth frames $\mathcal{Y}$, we first calculate their forward difference $\Delta \widehat{\mathcal{Y}}, \Delta \mathcal{Y} \in \mathbb{R}^{(T'-1) \times C \times H \times W}$, where:
\begin{equation}
\begin{aligned}
  \Delta \widehat{\mathcal{Y}}_i = \widehat{\mathcal{Y}}_{i+1} - \widehat{\mathcal{Y}}_{i}, \\
  \Delta \mathcal{Y}_i = \mathcal{Y}_{i+1} - \mathcal{Y}_{i}.
\end{aligned}
\end{equation}

Then, we transform the differences into probabilities by the softmax function on the channel, height, and width dimension and obtain $\sigma(\Delta \widehat{\mathcal{Y}}), \sigma(\Delta \mathcal{Y})$, where:
\begin{equation}
\begin{aligned}    
\sigma(\Delta \widehat{\mathcal{Y}})_{i,j,k,l} = \frac{\exp{(\Delta \widehat{\mathcal{Y}}_{i,j,k,l}} / \tau)}{\sum_{j'=1}^{C}\sum_{k'=1}^{H} \sum_{l'=1}^{W} \exp(\Delta \widehat{\mathcal{Y}}_{i, j',k',l'} / \tau)}, \\
\sigma(\Delta \mathcal{Y})_{i,j,k,l} = \frac{\exp{(\Delta \mathcal{Y}_{i,j,k,l}}) / \tau}{\sum_{j'=1}^{C}\sum_{k'=1}^{H} \sum_{l'=1}^{W} \exp(\Delta \mathcal{Y}_{i, j',k',l'} / \tau)},
\end{aligned}
\end{equation}
and $\tau$ represents the temperature parameter which we empirically set as 0.1 to sharpen the probability distribution. Through the competition mechanism of the softmax function~\cite{choromanski2020rethinking,peng2020random,pmlr-v162-wu22m}, the high difference frames are penalized.

Thus, the differential divergence regularization $\mathcal{L}_{reg}$ is defined as the Kullback-Leibler divergence between the probability distributions $\sigma(\Delta \widehat{\mathcal{Y}})$ and $\sigma(\Delta \mathcal{Y})$:
\begin{equation}
\begin{aligned}  
  \mathcal{L}_{reg}(\widehat{\mathcal{Y}}, \mathcal{Y}) &= D_{KL}(\sigma(\Delta \widehat{\mathcal{Y}}) \; || \; \sigma(\Delta \mathcal{Y}))\\
  &= \sum_{i=1}^{T'-1} \sigma(\Delta \widehat{\mathcal{Y}}_i) \log \frac{\sigma(\Delta \widehat{\mathcal{Y}}_i)}{\sigma(\Delta \mathcal{Y}_i)}.
\end{aligned}
\end{equation}

Our model is trained end-to-end in a fully unsupervised manner, and the overall objective function consists of the mean square error loss and the differential divergence regularization weighted by a constant $\alpha$:
\begin{equation}
  \mathcal{L} = \sum_{i=1}^{T'}\|\widehat{\mathcal{Y}} - \mathcal{Y} \|^2 + \alpha \mathcal{L}_{reg}(\widehat{\mathcal{Y}}, \mathcal{Y}),
\end{equation}
where the first term focuses on intra-frame-level differences, and the second regularization term focuses on inter-frame-level variations.

\section{Experiments} % 2-3 pages
In this section, we present experiments that demonstrate the effectiveness of our proposed method. The experiments are conducted on various datasets with different settings to validate our proposed model from three aspects:
\begin{itemize}
  \item Standard spatiotemporal predictive learning (Section~\ref{lab:standard}). We recognize the prediction problem of the same number of input and output frames as the standard spatiotemporal predictive learning. We evaluate the performance on standard spatiotemporal predictive learning and compare our model with state-of-the-art methods with Moving MNIST~\cite{srivastava2015unsupervised}and TaxiBJ~\cite{zhang2017deep} datasets.
  \item Generalization ability across different datasets (Section~\ref{lab:generalization}). Generalizing the learned knowledge to other domains is a challenge in unsupervised learning. We investigate the such ability of our method by training the model on the KITTI~\cite{geiger2013vision} dataset and evaluating it on the Caltech Pedestrian~\cite{dollar2009pedestrian} dataset. 
  \item Predicting frames with flexible lengths (Section~\ref{lab:flexible}). One of the advantages of recurrent units is that they can easily handle flexible-length frames like the KTH dataset~\cite{schuldt2004recognizing}. Our work tackles the long-length frame prediction by imitating recurrent units that feed predicted frames as the input and recursively produce long-term predictions. 
\end{itemize}

\subsection{Experimental Setups}

\paragraph{Datasets} We quantitatively evaluate our model on the following datasets for both synthetic and real-world scenarios: 
\begin{itemize}
  \item \textbf{Moving MNIST}~\cite{srivastava2015unsupervised} is a synthetic dataset consisting of two digits independently moving within the 64 $\times$ 64 grid and bouncing off the boundary. It is a standard benchmark in spatiotemporal predictive learning.
  \item \textbf{TaxiBJ} contains the trajectory data in Beijing collected from taxicab GPS with two channels, i.e., inflow or outflow defined in~\cite{zhang2017deep}. Following the previous works~\cite{mim,phydnet}, we normalize the data into $[0,1]$.
  \item \textbf{KTH}~\cite{schuldt2004recognizing} contains 25 individuals performing six types of actions. Following~\cite{villegas2017decomposing,e3dlstm}, we use person 1-16 for training and 17-25 for testing. Models are trained to predict the next 20 or 40 frames from the previous 10 observations.
  \item \textbf{Caltech Pedestrian} is a driving dataset focusing on detecting pedestrians. It consists of approximately 10 hours of $640\times 480$ videos taken from vehicles driving through regular traffic in an urban environment. We follow the same protocol of PredNet~\cite{prednet} and CrevNet~\cite{crevnet} for pre-processing, training, and evaluation.
\end{itemize}
We summarize the statistics of the above datasets in Table~\ref{tab:dataset}, including the number of training samples $N_{train}$ and the number of testing samples $N_{test}$.

\begin{table}[h]
\centering
\setlength{\tabcolsep}{1.4mm}{
\caption{The statistics of datasets. The training or testing set has $N_{train}$ or $N_{test}$ samples, composed by $T$ or $T'$ images with the shape $(C, H, W)$.}
\label{tab:dataset}
\begin{tabular}{cccccc}
\toprule
& $N_{train}$ & $N_{test}$ & $(C,H, W)$ & $T$ & $T'$ \\
\midrule
MMNIST    &  10000 &  10000 & (1, 64, 64)   & 10  & 10 \\
TaxiBJ &  19627 &  1334  & (2, 32, 32)   & 4   & 4 \\
KTH       &  5200  &  3167  & (1, 128, 128) & 10  & 20 or 40\\
Caltech & 2042   &1983 & (3, 128, 160) & 10 & 1\\
\bottomrule
\end{tabular}}
\end{table}

\paragraph{Measurement} Following~\cite{phydnet,crevnet}, we employ Mean Squared Error (MSE), Mean Absolute Error (MAE), Structure Similarity Index Measure (SSIM), and Peak Signal to Noise Ratio (PSNR) to evaluate the quality of predictions. MSE and MAE estimate the absolute pixel-wise errors, SSIM measures the similarity of structural information within the spatial neighborhoods, and PSNR is an expression for the ratio between the maximum possible power of a signal and the power of distorted noise.

\paragraph{Implementation details} We implement the proposed method with the Pytorch framework and conduct experiments on a single NVIDIA-V100 GPU. The model is trained with a mini-batch of 16 video sequences while the AdamW optimizer is utilized with a learning rate of 0.01 and a weight decay of 0.05.

\subsection{Standard spatiotemporal predictive learning}
\label{lab:standard}

\begin{figure*}[htbp]
\centering
\includegraphics[width=\textwidth]{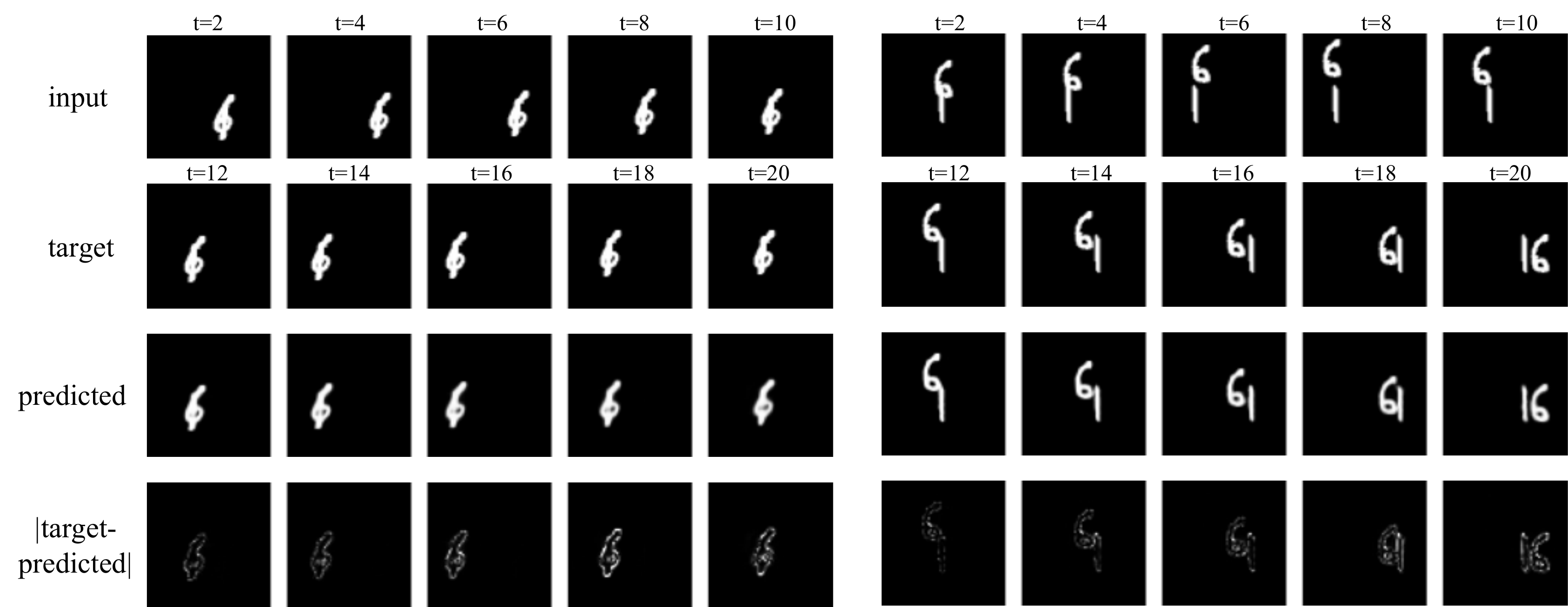} 
\caption{Qualitative visualization of predicted results on Moving MNIST dataset. The differences between the ground truth and the predicted frames are visualized in the last row.}
\label{fig:mmnist_example} 
\end{figure*}

\subsubsection{Moving MNIST} This dataset is a standard benchmark in spatiotemporal predictive learning. We evaluate our proposed method against strong recent baselines, including competitive recurrent architectures: ConvLSTM~\cite{convlstm}, PredRNN~\cite{predrnn}, PredRNN++~\cite{predrnn++}, MIM~\cite{mim}, LMC~\cite{lee2021video}, E3D-LSTM~\cite{e3dlstm}, Conv-TT-LSTM~\cite{su2020convolutional}, and CrevNet~\cite{crevnet}. We also compare the method DDPAE~\cite{ddpae} that is specifically designed for this dataset. The quantitative results are reported in Table~\ref{tab:mmnist}, and qualitative visualizations of the predicted results are shown in Figure~\ref{fig:mmnist_example}. 

Our proposed method significantly outperforms all the baselines above under three different metrics. The performance gain is large with respect to state-of-the-art recurrent methods.

\begin{table}[h]
\centering
\caption{Quantitative results of different methods on the Moving MNIST dataset ($10 \rightarrow 10$ frames).}
\setlength{\tabcolsep}{2.8mm}{
\begin{tabular}{c ccc}
\toprule
& \multicolumn{3}{c}{Moving MNIST} \\
Method & MSE$\downarrow$ & MAE$\downarrow$ & SSIM$\uparrow$ \\ 
\midrule
ConvLSTM~\cite{convlstm}    & 103.3 & 182.9 & 0.707 \\
VPN~\cite{kalchbrenner2017video} & 64.1 & - & 0.870 \\
PredRNN~\cite{predrnn}      & 56.8  & 126.1 & 0.867 \\
PredRNN++~\cite{predrnn++}  & 46.5  & 106.8 & 0.898 \\
MIM~\cite{mim}              & 44.2  & 101.1 & 0.910 \\
LMC~\cite{lee2021video}     & 41.5  & -     & 0.924 \\
E3D-LSTM~\cite{e3dlstm}     & 41.3  & 87.2  & 0.910 \\
Conv-TT-LSTM~\cite{su2020convolutional} & 53.0 & - & 0.915 \\
DDPAE~\cite{ddpae}          & 38.9  & 90.7  & 0.922 \\
PhyDNet~\cite{phydnet}      & 24.4  & 70.3  & 0.947 \\ 
SimVP~\cite{Gao_2022_CVPR}  & 23.8  & 68.9 & 0.948 \\
Crevnet~\cite{crevnet}      & 22.3  & -     & 0.949 \\
\hline
Ours       & \textbf{19.8}  & \textbf{60.3} & \textbf{0.957} \\  
\bottomrule
\end{tabular}}
\label{tab:mmnist}
\end{table}

\subsubsection{TaxiBJ} We evaluate our proposed model on a complicated real-world dataset, TaxiBJ~\cite{zhang2017deep}. Driven by human consciousness, the complex real-world traffic flows requires modeling transport phenomena and traffic diffusion for prediction. Due to the spatiotemporal nature of the traffic forecasting task, we straightforwardly implement our model for it. 

\begin{figure}[htbp]
\centering
\includegraphics[width=0.48\textwidth]{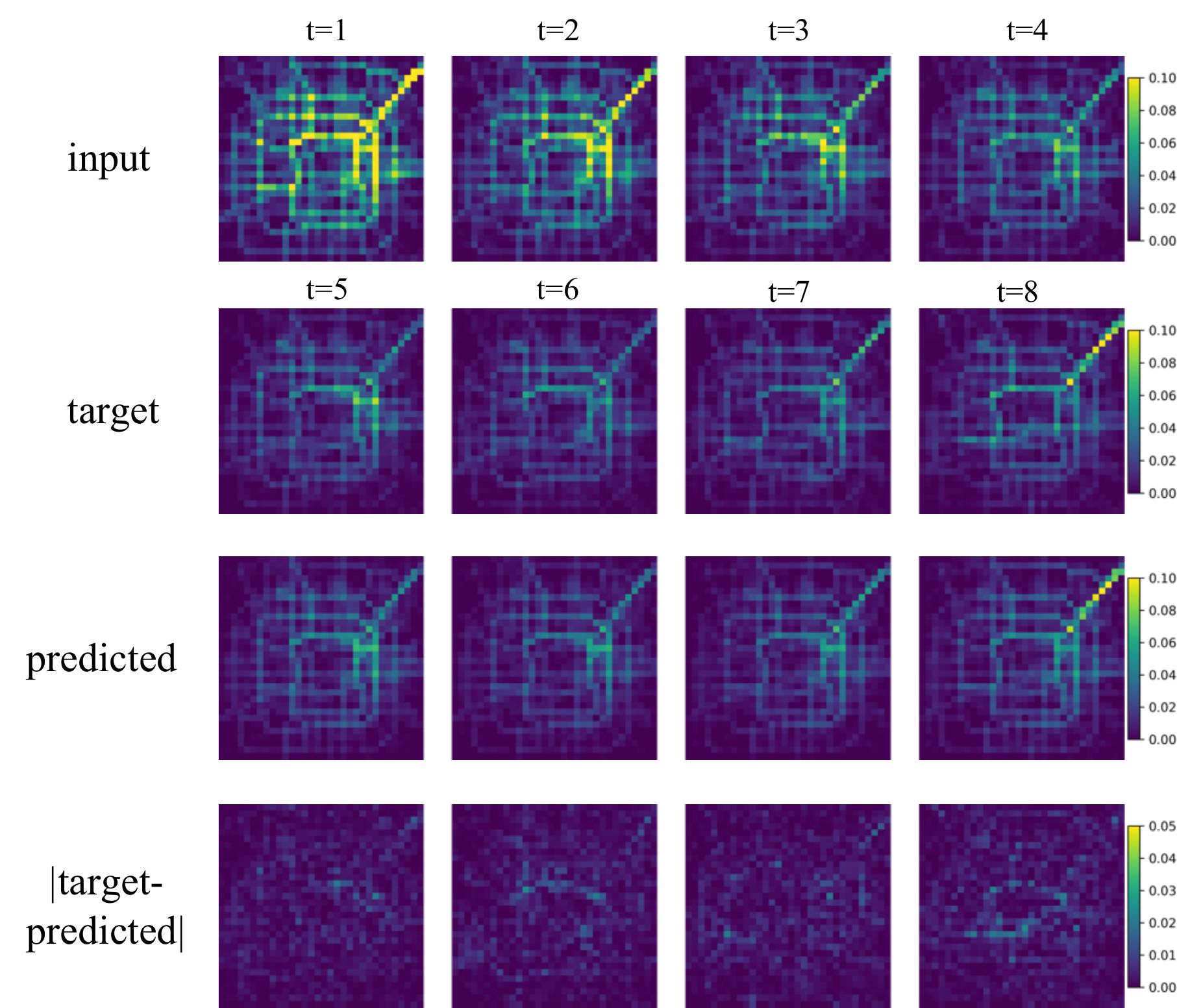} 
\caption{Qualitative visualization of predicted results on TaxiBJ dataset. The differences between the ground truth and the predicted frames are visualized in the last row.}
\label{fig:traffic_example} 
\end{figure}

The qualitative visualizations of the predicted results are shown in Figure~\ref{fig:traffic_example}, and the quantitative results are reported in Table~\ref{tab:traffic}. Though the given frames are quite different from the future frames, our model can still accurately produce reliable frames. The difference between target frames and predicted frames is mainly located in central spots, but the overall trend is approximating the ground truth. It can also be observed that the quantitative results are consistently well, which demonstrates the practical application value in real-world scenarios.

\begin{table}[h]
\centering
\caption{Quantitative results of different methods on the TaxiBJ dataset ($4 \rightarrow 4$ frames).}
\setlength{\tabcolsep}{2.5mm}{
\begin{tabular}{c ccc}
\toprule
& \multicolumn{3}{c}{TaxiBJ} \\
Method & MSE $\times$ 100$\downarrow$ & MAE$\downarrow$ & SSIM$\uparrow$ \\ 
\midrule
ConvLSTM~\cite{convlstm}    & 48.5  & 17.7 & 0.978 \\
PredRNN~\cite{predrnn}      & 46.4  & 17.1 & 0.971 \\
PredRNN++~\cite{predrnn++}  & 44.8  & 16.9 & 0.977 \\
MIM~\cite{mim}              & 42.9  & 16.6 & 0.971 \\
E3D-LSTM~\cite{e3dlstm}     & 43.2  & 16.9 & 0.979 \\
PhyDNet~\cite{phydnet}      & 41.9  & 16.2 & 0.982 \\ 
SimVP~\cite{Gao_2022_CVPR}  & 41.4  & 16.2 & 0.982 \\

\hline
Ours       & \textbf{34.4}  & \textbf{15.6} & \textbf{0.983} \\  
\bottomrule
\end{tabular}}
\label{tab:traffic}
\end{table}

\subsection{Generalization across different datasets}
\label{lab:generalization}

The generalization ability is one of the fundamental problems in artificial intelligence technology. Traditional supervised learning suffers from its poor generalization of labeled datasets with different domains. Self-supervised learning aims to learn robust representations based on unlabeled data and evaluates the generalization ability of the learned model. While contrastive self-supervised learning and masked self-supervised learning in visual tasks usually evaluate such generalization ability by downstream tasks, we evaluate it by the prediction results across different datasets in spatiotemporal predictive learning. 

\begin{figure}[htbp]
\centering
\includegraphics[width=0.48\textwidth]{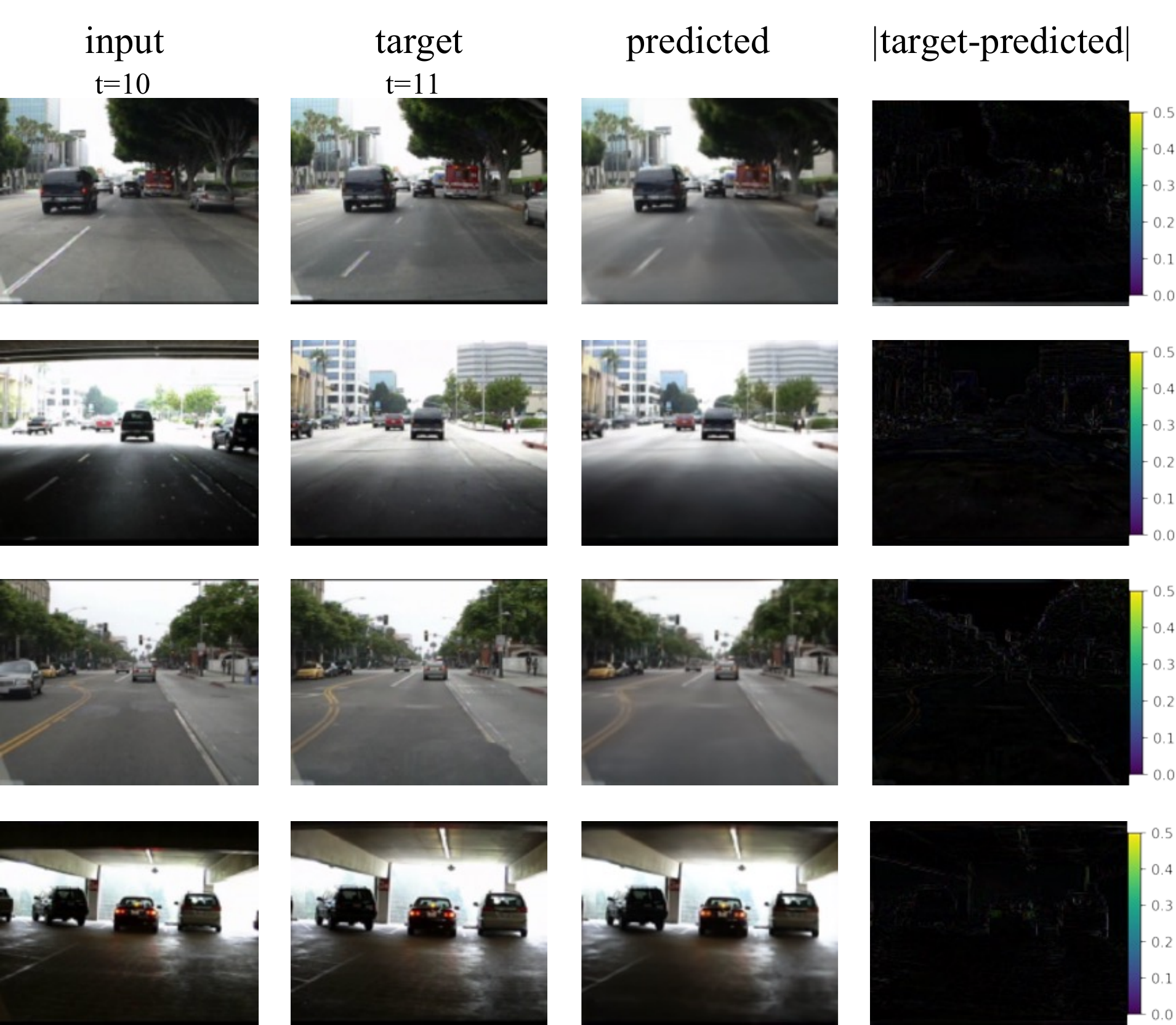} 
\caption{Qualitative visualization of predicted results on Caltech dataset. The differences between the ground truth and the predicted frames are visualized in the last row.}
\label{fig:caltech_example} 
\end{figure}

Following the previous works~\cite{prednet, crevnet, liang2017dual}, we train the model using the raw video sequences from the KITTI dataset and evaluate the model by Caltech Pedestrian dataset that is made to match the frame rate and image size ($128 \times 160$) of the KTTI dataset. The final prediction is made for the next frame after a 10-frame warm-up.

We show the qualitative visualization in Figure~\ref{fig:caltech_example} and report the quantitative results in Table~\ref{tab:caltech}. Note that some of the baseline results are copied from~\cite{oprea2020review}. It can be seen that our proposed method achieves state-of-the-art performance under both SSIM and PSNR metrics in this generalization evaluation task. Moreover, our model shows significantly robust predictions in terms of the variation of illumination and lane line, suggesting its practical applications in autonomous vehicles.

\begin{table}[h]
\centering
\caption{Quantitative results of different methods on the Caltech Pedestrian dataset ($10 \rightarrow 1$ frame).}
\setlength{\tabcolsep}{6mm}{
\begin{tabular}{c cc}
\toprule
& \multicolumn{2}{c}{Caltech Pedestrian} \\
Method & SSIM$\uparrow$ & PSNR$\uparrow$ \\ 
\midrule
BeyondMSE~\cite{mathieu2015deep}    & 0.847 & -    \\
MCnet~\cite{villegas2017decomposing}& 0.879 & -    \\
DVF~\cite{liu2017video}             & 0.897 & 26.2 \\
Dual-GAN~\cite{liang2017dual}       & 0.899 & -    \\
CtrlGen~\cite{hao2018controllable}  & 0.900 & 26.5 \\
PredNet~\cite{prednet}              & 0.905 & 27.6 \\
ContextVP~\cite{byeon2018contextvp} & 0.921 & 28.7 \\
GAN-VGG~\cite{shouno2020photo}      & 0.916 & -    \\
G-VGG~\cite{shouno2020photo}        & 0.917 & -    \\
SDC-Net~\cite{reda2018sdc}          & 0.918 & -    \\
rCycleGan~\cite{kwon2019predicting} & 0.919 & 29.2 \\
DPG~\cite{gao2019disentangling}     & 0.923 & 28.2 \\
G-MAE~\cite{shouno2020photo}        & 0.923 & -    \\
GAN-MAE~\cite{shouno2020photo}      & 0.923 & -    \\
CrevNet~\cite{crevnet}              & 0.925 & 29.3 \\
STMFANet~\cite{jin2020exploring}    & 0.927 & 29.1 \\
SimVP~\cite{Gao_2022_CVPR}          & 0.940 & 33.1 \\
\hline
Ours       & \textbf{0.946}  & \textbf{33.7} \\  
\bottomrule
\end{tabular}}
\vspace{-3mm}
\label{tab:caltech}
\end{table}

\subsection{Predicting frames with flexible lengths}
\label{lab:flexible}

Though recurrent units are adept at handling flexible-length frames, our model can also easily tackle such problems by imitating recurrent units that feed predicted frames as the input and recursively produce predictions. For this KTH dataset, our model is trained to predict the next 20 or 40 frames from the given 10 observations. Moreover, this dataset contains six types of human actions (walking, jogging, running, boxing, hand waving and hand clapping) performed several times by 25 subjects in four different scenarios: outdoors, outdoors with scale variations, outdoors with different clothes and indoors. The difficulty of this human motion prediction task not only lies in its flexible lengths of predicted frames but also in its complex dynamics involving the randomness of human consciousness. Following~\cite{predrnn}, we use the Peak Signal Noise Ratio (PSNR) and the Structural Similarity Index Measure (SSIM) as evaluation metrics to measure the framewise prediction quality from the perceptive view. 
The detailed quantitative results are shown in Table~\ref{tab:kth}.

\begin{table}[h]
\centering
\caption{Quantitative results of different methods on the KTH dataset ($10 \rightarrow 20$ and $10 \rightarrow 40$ frames).}
\setlength{\tabcolsep}{1.5mm}{
\begin{tabular}{ccccc}
\toprule
& \multicolumn{2}{c}{KTH ($10 \rightarrow 20$)} & \multicolumn{2}{c}{KTH ($10 \rightarrow 40$)} \\ \cline{2-5} 
Method  & SSIM$\uparrow$ & PSNR$\uparrow$ & SSIM$\uparrow$ & PSNR$\uparrow$ \\
\midrule
MCnet~\cite{villegas2017decomposing}  & 0.804 & 25.95 & 0.73 & 23.89 \\
ConvLSTM~\cite{convlstm} & 0.712 & 23.58 & 0.639 & 22.85 \\
SAVP~\cite{lee2018stochastic} & 0.746 & 25.38 & 0.701 & 23.97 \\
VPN~\cite{kalchbrenner2017video} & 0.746 & 23.76 & --  & --  \\
DFN~\cite{jia2016dynamic} & 0.794 & 27.26 & 0.652 & 23.01 \\
fRNN~\cite{oliu2018folded} & 0.771 & 26.12 & 0.678 & 23.77 \\
Znet~\cite{zhang2019z} & 0.817 & 27.58 & --    & --    \\
SV2Pv~\cite{babaeizadeh2018stochastic} & 0.838 & 27.79 & 0.789 & 26.12 \\
PredRNN~\cite{predrnn}      & 0.839 & 27.55 & 0.703 & 24.16 \\
VarNet~\cite{jin2018varnet} & 0.843 & 28.48 & 0.739 & 25.37 \\
SAVP-VAE~\cite{lee2018stochastic} & 0.852 & 27.77 & 0.811 & 26.18 \\
PredRNN++~\cite{predrnn++} & 0.865 & 28.47 & 0.741 & 25.21 \\
MSNET~\cite{lee2018mutual} & 0.876 & 27.08 & --    & --    \\
E3d-LSTM~\cite{e3dlstm}    & 0.879 & 29.31 & 0.810 & 27.24 \\
STMFANet~\cite{jin2020exploring} & 0.893 & 29.85 & 0.851 & 27.56 \\ 
SimVP~\cite{Gao_2022_CVPR} & 0.905 & 33.72 & 0.886 & 32.93 \\
\midrule
Ours & \textbf{0.911} & \textbf{34.13} & \textbf{0.897} & \textbf{33.01} \\
\bottomrule
\end{tabular}}
\label{tab:kth}
\vspace{-4mm}
\end{table}

\subsection{Empirical Running Time}
TAU benefits from the parallelizable computation architecture, which leads to fast convergence and high training speed. We evaluate our efficiency by measuring the running time against state-of-the-art spatiotemporal predictive learning methods.

\begin{figure}[htbp]
\vspace{-2mm}
\centering
\includegraphics[width=0.42\textwidth]{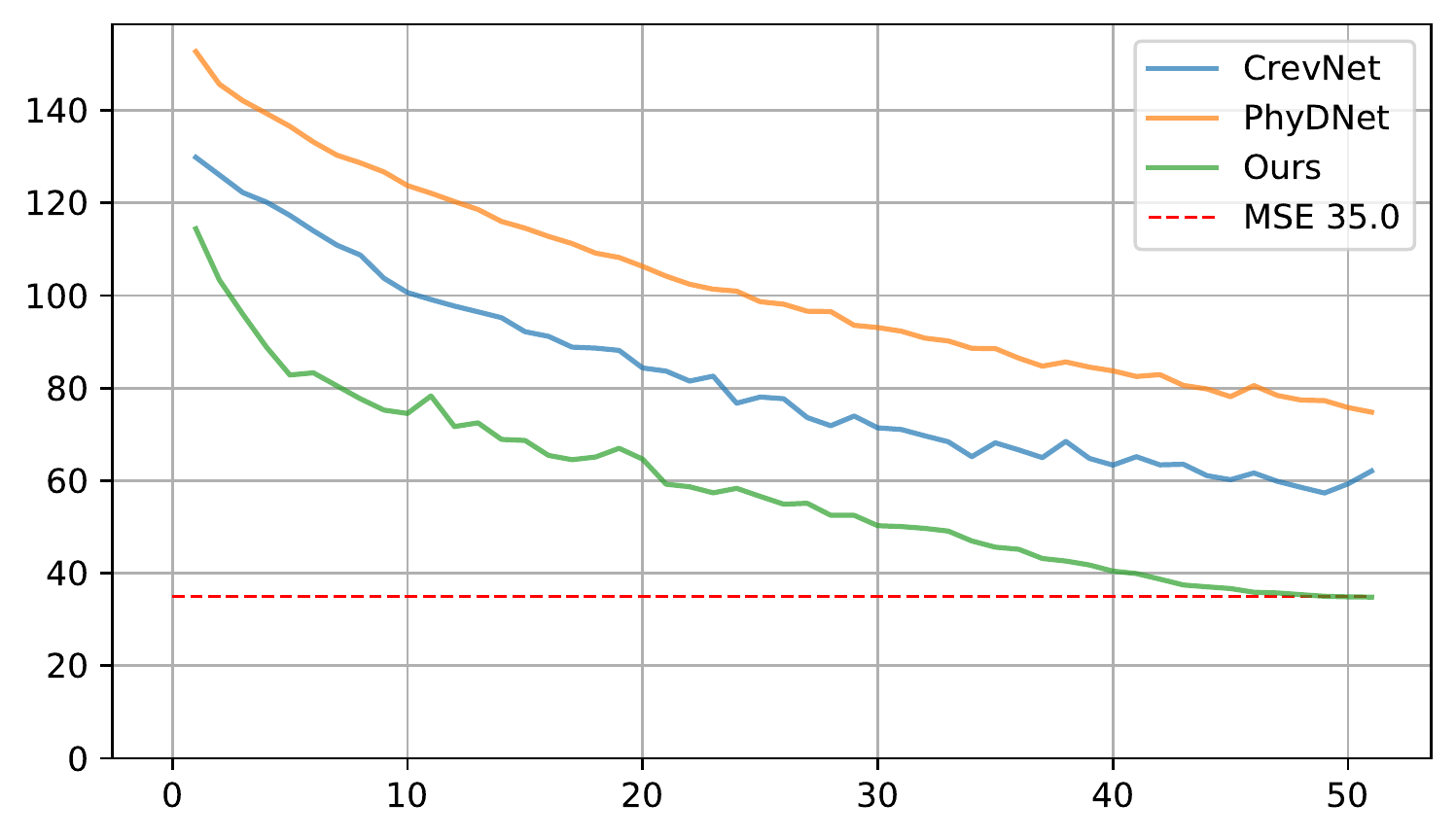} 
\caption{The learning curve comparison between state-of-the-art methods and ours (evaluated by MSE). The red dotted line denotes MSE 35.0, and only the first 50 epochs are shown.}
\vspace{-2mm}
\label{fig:cmp_mmnist} 
\end{figure}

The experiments are conducted on a single Tesla V100 GPU, and the batch size is set as 16. For the Moving MNIST dataset, CrevNet~\cite{crevnet} needs about 30 minutes per epoch, and PhyDNet~\cite{phydnet} needs about 7 minutes per epoch. Our model only requires 2.5 minutes per epoch. Furthermore, our method is able to converge at a rapid rate. As shown in Figure~\ref{fig:cmp_mmnist}, on the Moving MNIST dataset, our model can achieve MSE 35.0 with only 50 epochs, while CrevNet and PhyDNet are far from this performance. 

\subsection{Computational Cost and Ablation Study}

We compare the performance and computational cost with state-of-the-art methods in the first several rows in Table~\ref{tab:ablation}. Our model achieves superior results with better performance and much lower Flops. We also conduct ablation studies and summarize the results in Table~\ref{tab:ablation}. It can be seen that replacing TAU with the same number of convolutional blocks with vanilla $3\times 3$ convolutions (Conv Baseline) significantly degrades the performance. Training our model without differential divergence regularization (w/o DDR) will also weaken the prediction results. Both the TAU module and differential divergence regularization are useful. We also find that SA and DA of the TAU module play important roles in better performance.

\begin{table}[h]
\centering
\caption{Ablation study of our proposed method.}
\setlength{\tabcolsep}{1.2mm}{
\begin{tabular}{cccc}
\toprule
Method & {M-MNIST} & {TaxiBJ ($\times 100$)} & Flops(G) \\
\midrule
PredRNN        & 56.8  & 46.4 & 115.94 \\
PredRNN++      & 46.4  & 44.8 & 171.73 \\
MIM            & 44.2  & 42.9 & 179.17 \\
E3DLSTM        & 41.3  & 43.2 & 298.85 \\
SimVP          & 23.8  & 41.4 & 19.43  \\
\midrule       
Conv Baseline  & 58.9  & 43.5 & 6.11   \\ 
Ours w/o SA    & 23.2  & 40.8 & 15.30 \\
Ours w/o DA    & 22.4  & 38.4 & 15.96 \\
Ours w/o DDR   & 21.1  & 37.7 & 15.96  \\
% \midrule
Ours & \textbf{19.8} & \textbf{34.4} & 15.96 \\
\bottomrule
\end{tabular}
}
\vspace{-4mm}
\label{tab:ablation}
\end{table}

\section{Conclusion}

In this paper, we present a general framework of spatiotemporal predictive learning and propose an attention-based temporal module to replace the common-used recurrent units. By decomposing the temporal module into the intra-frame statical attention and the inter-frame dynamical attention, our proposed TAU module can achieve competitive performance across various experimental settings and datasets. Moreover, a novel differential divergence regularization is proposed to overcome the drawback of MSE loss that only considers the intra-frame error. In summary, our work highlights the importance of both intra-frame and inter-frame variations that enable the model to capture long-term relations and provide a new paradigm of efficient spatiotemporal predictive learning.

\paragraph{Acknowledgement.}
\vspace{-2mm}
We thank the anonymous reviewers for their constructive and helpful reviews. This work was supported by the National Key R\&D Program of China (2022ZD0115100), the National Natural Science Foundation of China (U21A20427), the Competitive Research Fund (WU2022A009) from the Westlake Center for Synthetic Biology and Integrated Bioengineering

% \paragraph{Acknowledgemnet}
% This work is supported in part by the Science and Technology Innovation 2030 - Major Project (No. 2021ZD0150100) and National Natural Science Foun- dation of China (No. U21A20427).

% \newpage
%%%%%%%%% REFERENCES
{\small
\bibliographystyle{ieee_fullname}
\bibliography{ref}
}

\end{document}